\documentclass{sig-alternate-2013}

\usepackage{times}
\usepackage{tikz}
\usepackage{pgf}
\usetikzlibrary{arrows,automata}
\usepackage{balance}

\usepackage{url}
\usepackage{xspace}
\usepackage{color}
\usepackage{amsmath} 
\usepackage{amssymb}
\usepackage{ifthen}
\usepackage{graphicx}
\usepackage{multirow}
\usepackage[T1]{fontenc}
\usepackage[utf8]{inputenc}
\usepackage{tabularx,ragged2e,booktabs,caption}
\newcolumntype{C}[1]{>{\Centering}m{#1}}

\newtheorem{thm}{Theorem}
\newtheorem{defn}[thm]{Definition}
\newtheorem{lemma}[thm]{Lemma}

\newcommand{\para}[1]{\par \smallskip \noindent {\bf #1.}}

\newcommand{\CX}{{\cal X}}
\newcommand{\tmix}{t_{\mathrm{mix}}}
\newcommand{\LAMP}{\mathsf{lamp}}
\newcommand{\GLAMP}{\mathsf{glamp}}
\newcommand{\MC}{\mathsf{MC}}
\newcommand{\tuple}[1]{\left\langle #1 \right\rangle}
\newcommand{\reals}{\mathbb{R}}
\newcommand{\nnz}{\mathrm{nnz}}

\def\sneq{{\tiny \mbox{$\neq$}}}
\def\subsetneq{\ \lower.5ex\hbox{$\stackrel{\subset}{\small \sneq}$}\ }

\newcommand{\omt}[1]{}

\newcommand{\graf}[3]{
\ifthenelse
 {\equal{#1}{h}}
  {\includegraphics[height=#2\textheight]{#3}}
  {\includegraphics[width=#2\textwidth]{#3}}
}

\usepackage{array}
\newcolumntype{C}[1]{>{\centering}m{#1}}

\permission{\copyright 2017 International World Wide Web Conference Committee \\ (IW3C2), published under Creative Commons CC BY 4.0 License.}
\conferenceinfo{WWW 2017,}{April 3--7, 2017, Perth, Australia.}
\copyrightetc{ACM \the\acmcopyr}
\crdata{978-1-4503-4913-0/17/04. \\
http://dx.doi.org/10.1145/3038912.3052644 \\
\includegraphics{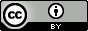}
}

\makeatletter
\def\thebibliography#1{%
\ifnum\addauflag=0\addauthorsection\global\addauflag=1\fi
     \section[References]{
        {References} 
          {\vskip -2pt plus 1pt} 
         \@mkboth{{\refname}}{{\refname}}%
     }%
     \list{[\arabic{enumi}]}{%
         \settowidth\labelwidth{[#1]}%
         \leftmargin\labelwidth
         \advance\leftmargin\labelsep
         \advance\leftmargin\bibindent
         \parsep=0pt\itemsep=1pt 
         \itemindent -\bibindent
         \listparindent \itemindent
         \usecounter{enumi}
     }%
     \let\newblock\@empty
     \raggedright 
     \sloppy
     \sfcode`\.=1000\relax
}
\makeatother

\begin{document} 


\title{Linear Additive Markov Processes}

\numberofauthors{4}

\author{
Ravi Kumar 
\hspace*{1cm}
Maithra Raghu\thanks{On leave from Cornell University.}
\hspace*{1cm}
Tam\'{a}s Sarl\'{o}s
\hspace*{1cm}
Andrew Tomkins \\
       \affaddr{Google}\\
       \affaddr{Mountain View, CA}\\
       \email{\{ravi.k53, maithrar, stamas, atomkins\}@gmail.com}
} 
\date{\today}

\maketitle

\begin{abstract}
  We introduce LAMP: the Linear Additive Markov Process.  Transitions
  in LAMP may be influenced by states visited in the distant history
  of the process, but unlike higher-order Markov processes, LAMP
  retains an efficient parameterization.  LAMP also allows the
  specific dependence on history to be learned efficiently from data.

  We characterize some theoretical properties of LAMP, including its
  steady-state and mixing time.  We then give an algorithm based on
  alternating minimization to learn LAMP models from data.
   
  Finally, we perform a series of real-world experiments to show that
  LAMP is more powerful than first-order Markov processes, and even
  holds its own against deep sequential models (LSTMs) with a
  negligible increase in parameter complexity.
\end{abstract}

\section{Introduction}

Markov processes~\cite{karlin-stochastic} are arguably the most widely
used and broadly known sequence modeling technique available.  They
are simple, elegant, and theoretically well understood.  They have
been extended in many directions, and form a useful building block for
many more sophisticated approaches in machine learning and data
mining.  However, the Markovian assumption that transitions are
\emph{independent} of history given the current state is routinely
violated in real-world data sequences, as the user's full history
provides powerful clues about the likely next state.

To address these longer-range dependencies, Markov processes have been
extended from first-order models to $k$th-order models in which the
next element depends on the previous $k$.  Such models, however, have
a parameter space that grows exponentially in $k$, so for large state
spaces, even storing the most recent 5--6 elements of history may
encounter data sparsity issues and impact generalization.
Variable-order Markov processes~\cite{vomc2} attempt to address these
issues by retaining variable amounts of history, allocating more
resources to remembering more important recent situations.  Even here,
a dependence on an element from the past can be used only if all
intermediate sequences of states are remembered explicitly; there are
exponentially many such sequences, and they may not impact the
prediction in any way.

Beyond Markov processes, a wide range of more complex sequence models
have been proposed.  The last two decades have seen notable empirical
success for models based on deep neural networks, particularly the
\emph{Long Short-Term Memory (LSTM)} architecture~\cite{1997lstm}.
These LSTM models show state of the art performance in many domains,
at the cost of low interpretability, long training times, and the
requirement for significant parameter tuning.

\para{Basic intuition}
Our goal is to propose a model with the simplicity, parsimony, and
interpretability of first-order Markov processes, but with ready
access to possibly distant elements of history.  The family of models
we propose also has clean mathematical behavior, is easy to interpret,
and comes equipped with an efficient learning algorithm.


The intuition behind our model is as follows.  We select a sampled
point from the user's history, according to a learned time decay.  We
then assume that the user will adopt the same behavior as somebody in
that prior context.  Thus, if a recency-weighted sample of the user's
past shows significant time spent in the ``parent'' state and the
``office worker'' state, the model will offer next state predictions
based on the current state, mixed with predictions that are
appropriate for ``parent'' and ``office worker.''  The relative
importance of the current state versus the historical norms will be
learned from the data.

\para{An overview of LAMP}
Based on this intuition, we define a Linearly Additive Markov Process,
or LAMP, based on a stochastic matrix $P$, and a distribution $w$ over
positive integers.  The matrix determines transitions from a given
context, and the distribution captures how far back into history the
process will look to select a context for the user.  Given a user who
has traversed a sequence of states $X=x_1x_2\ldots x_{t}$, the process
first selects a past state $x_{i}$ with probability proportional to
$w_{t-i+1}$, and then selects a next state $y$ using $P$ as a
transition from $x_i$.  Thus, the user chooses a next step by taking a
step forward from some state chosen from the history.  The matrix $P$
and the distribution $w$ are to be learned from data.

One may view LAMP as factoring the temporal effect from the contextual
effect, resulting in a multiplicative model.  When a state is chosen
from the history, the behavior conditioned on that state is
independent of how long ago the state was visited.  This factoring
keeps the model parsimonious.

We also consider a natural generalization of LAMP which we refer to as
Generalized LAMP, or GLAMP.  This extended model allows us to
associate different matrices with different regions of history.  For
example, one matrix might capture transitions from the current state,
taking into account geographic locality.  Another matrix might capture
transitions from past states, taking into account co-visitations.  We
define and characterize GLAMP.

\para{Theoretical properties}
LAMP is a generative sequence model related to Markov processes, but
more expressive.  It is therefore critical for us to understand the
mathematical properties of the process, so we can evaluate whether
LAMP is appropriate to model a particular dataset.

With this motivation in mind, we study the theoretical properties of
LAMP.  We show that LAMP models with $k$ elements of history cannot be
approximated within any constant factor by Markov processes of order
$k-1$, but can be captured completely by Markov processes of order $k$
(albeit with a parameterization that is exponentially larger in $k$).
We show further that LAMP models under some mild regularity conditions
do possess a limiting equilibrium distribution over states, and this
distribution is equal to the steady state of the first-order Markov
process defined by matrix $P$.  However, beyond the steady state
itself, the transition dynamics of the LAMP process are quite
different from that of the Markov process on the same matrix, and are
in general different from any first-order Markov Process.  Finally,
again under some mild conditions, we are able to characterize the
convergence of LAMP to its steady state, in terms of properties of $P$
and $w$.  In order to prove these results, we develop an alternate
characterization of the process, dubbed the \emph{exponent process},
which allows us to invoke the machinery of renewal processes in the
analysis.

\para{Learning and evaluation}
Given data, the challenge for LAMP is to determine which stochastic
matrix and distribution over positive integers will maximize
likelihood of the observations.  The learned matrix could in general
be different from the empirical transition matrix, as the model might
choose to ``explain'' certain transitions by reference to historical
states rather than the current state.

To perform the learning, we show that the likelihood is convex in
either $P$ or $w$ alone, and we give an alternating maximization
procedure.  We will discuss the optimization issues that arise in
fitting this model accurately.



We then perform a series of experiments to evaluate the performance of
LAMP.  Our expectation is that LAMP will be useful in the same
circumstances as Markov processes: as simple, interpretable, and
extensible mathematical models, or as modules in larger statistical
systems.  We therefore begin by comparing LAMP to $k$th-order Markov
processes, with and without smoothing.  Of the four datasets we
consider, our results show that for two of them, there is significant
benefit to having the additional history provided by LAMP, and
perplexity may be reduced by 20--40\%.  For the other two datasets we
consider, the LAMP optimizer sets the weight distribution to place all
mass on the most recent state, causing LAMP to fall back to a
first-order Markov process.  We therefore conclude that when benefits
are attainable through long-distance dependencies, LAMP may show
significant gains; otherwise, it degrades gracefully to a first-order
Markov process.  In this comparison LAMP and first-order Markov
processes have similar parameter complexity: a large transition matrix
in both cases, and optionally $k$ additional scalars for LAMP.  To see
any lift with such a modest additional parameterization is
interesting.  For completeness, we also compare LAMP with LSTMs. Our
experiments show an intriguing outcome: neither LSTM nor LAMP is a
clear winner on all datasets.

%

Overall, we find that LAMP represents an intriguing and valuable new
point on the curve of accuracy versus complexity in finite-state
sequence modeling.

\section{Related Work}
\label{sec:rel}

Markov processes have been studied for more than a century.  The line
of work most closely related to ours is that of Additive Markov
processes, in which the Markovian assumption is relaxed to take some
history into account.  In particular, the probability of the next
state is a sum of memory functions, where each function depends on the
next state and the historical state.  They were formally introduced as
a probabilistic process by
A. A. Markov~\cite{aamarkov1971additivemarkov}, where versions of
existing limit theorems for Markov processes were extended to this new
framework.  A first usage of these models to represent long range
dependencies came in the context of physics and dynamical
systems~\cite{usatenko2003symbolic, melnyk2006memory,
  usatenko2009random}.  In all of these works, however, the state
space is restricted to binary and the construction of the memory
function is empirical.

Independent of our work and recently, Zhang et
al.~\cite{zhang2014lore, zhang2015spatiotemporal} use a ``gravity
model'', which is similar to LAMP in that it is parameterized by a
transition matrix and a history distribution.  However, they make no
attempt to learn the parameters in a robust manner.  Instead, the
transition matrix is estimated as the empirical transition matrix of a
first-order Markov process from the data, and the history distribution
is assumed to be from a particular family. We, in addition to
providing a theoretical treatment of LAMP, also develop a principled
method to learn the parameters without assumptions on the form of the
history function.

There have been several lines of work that go beyond the Markovian
assumption in order to capture user behavior in real-world settings.
Higher-order and variable-order Markov
processes~\cite{vomc2,vomc,rissanen} use the last few states in the
history to determine the probability of transition to the next state.
While they are strictly more powerful than the first-order Markov
process, they suffer from two drawbacks.  First, they are expensive to
represent since the amount of space needed to store the transition
matrix could be exponential in the order, and second, from a learning
point of view they are particularly vulnerable to data sparsity.
LAMP, on the other hand, is succinctly representable and has no data
sparsity issues.

Motivated by web browsers, Fagin et al.~\cite{backbutton} propose the
`Back Button' random walk in which users can revisit steps immediately
prior in their history as the next state they transition to.  However
the back button model erases history when moving backwards, while LAMP
retains it.  Pemantle et al.~\cite{pemantle1992vertex} suggest a
node-reinforced random walk, where popular nodes are re-visited with
higher proportions. This is one method for utilizing user history
while maintaining a small set of parameters. But the total number of
visits to a node cannot be so easily adapted to take into account
temporal relevance.  LAMP, on the other hand, can easily incorporate
recency via the history distribution.  This notion of using summary
statistics is also studied in Kumar et al.~\cite{kumar2015inverting}
in a different framework, with summary statistics used to infer facts
about an underlying generative process, which is assumed to be
Markovian.

Markov processes have been used in several web mining settings to
model user behavior.  PageRank~\cite{PR} is a classic use of random
walks to capture web surfing behavior.  First and higher-order Markov
processes have been used to model user browsing patterns~\cite{PR,
  Ramesh,Sen, Zukerman, Pirolli} and variable-order Markov processes
have been used to model session logs~\cite{Borges}.  First-order
Markov processes~\cite{Boldi, Craswell} and variable-order hidden
Markov processes~\cite{Cao} have often been applied to context-aware
search, document re-ranking, and query suggestions.  Chierichetti et
al.~\cite{chierichetti2012web} study the problem of using
variable-order Markov processes to model user behavior in different
online settings. Singer et al.~\cite{singer2015hyptrails} combine
first-order Markov processes with Bayesian inference for comparing
different prior beliefs about transitions between states and selecting
the Dirichlet prior best supported by data.  Our experiments suggest
that LAMP can go well beyond first-order Markov processes, with
negligible increase in space and learning complexities.  We believe
LAMP will be a powerful way to model user behavior for many of these
applications.

\newcommand{\mymath}[1]{$$ #1 $$}

\section{LAMP}
\label{sec:model}

In this section we introduce the Linear Additive Markov Process
(LAMP), and describe its properties.  

\subsection{Background}

Let $[n] = \{ 1, 2, \ldots, n \}$.  Let $\CX$ be a state space of size
$n$.  We use lower case $x, y, \ldots$ and its subscripted versions to
denote elements of $\CX$ and upper case $X, Y, \ldots$ and its
subscripted versions to denote random variables defined on $\CX$.  Let
$\{ P(x, y) \}$ be an $n \times n$ stochastic matrix defined on $\CX$,
with $P(x, y) \geq 0$ and for all $x \in \CX$, $\sum_{y \in \CX} P(x,
y) = 1$.

The matrix $P$ naturally corresponds to a first-order Markov process
on the state space $\CX$ in which, given a current state $x$, the
process jumps to state $y$ with probability $P(x, y)$.  Formally, if
$x_0, \ldots, x_{t-1}$ are the states in the process so far, then
\begin{equation}
\label{eqn:mc}
\Pr[X_t = x \mid x_0, \ldots, x_{t-1}] = P(x_{t-1}, x).
\end{equation}
Let $P^t$ denote the matrix raised to the power $t$.  Let $\nnz(P)$
denote the number of non-zero entries of $P$.

Let $\pi$ be the stationary distribution of $P$ satisfying
\mymath{
\sum_x P(x, y) \cdot \pi(x) = \pi(y), \mbox{ for all } y \in \CX.
}
Equivalently, $\pi = \lim_{t \rightarrow \infty} \pi_0 \cdot P^t$ for
any starting distribution $\pi_0$.  If $P$ is ergodic (see,
e.g.,~\cite{karlin-stochastic} for formal definitions), then the
stationary distribution $\pi$ of $P$ is well-defined and unique.

While the stationary distribution captures the asymptotic behavior of
the Markov process, it does not quantify how fast it approaches this
limit.  The notion of mixing time captures this: it is the number of
steps needed before the Markov process converges to its stationary
distribution, regardless of the node from where it started.  Formally,
let $1_z$ denote the unit vector with $1$ in the $z$th position, and
let
\mymath{
\Delta(P, t, z) = \frac{1}{2} \left|\pi - 1_z \cdot P^t \right| =
\frac{1}{2} \sum_{x} \left|\pi(x) - (1_z \cdot P^t)(x) \right|,
}
denote the \emph{total variation distance} between the stationary
distribution and the distribution after $t$ steps when the Markov
process starts at $z$.  The \emph{mixing time}, with respect to a
parameter $\delta > 0$, is then defined as:
\mymath{
\tmix(P, \delta) = 
 \max_z \left\{  \min_t \left\{ \Delta(P, t', z) \leq \delta, \mbox{ for all } t'
 \geq t \right\} \right\}. 
}

A \emph{$k$th-order Markov process, $\MC_k$}, for $k > 1$, is defined as
follows: let $P_k: \CX^{k} \times \CX \rightarrow \reals$ be a stochastic
matrix on $\CX^k$, the space of $k$-tuples on $\CX$.  Then,
\begin{equation}
\label{eqn:kmc}
\Pr[X_t = x_t \mid x_0, \ldots, x_{t-1}] = P(\tuple{x_{t-k}, \ldots, x_{t-1}}, x_t).
\end{equation}
where $t > k$;  we assume there is a starting distribution
on tuples of length $k-1$.

Let $k$ be a non-negative integer.  Let $w = (w_1, \ldots, w_k)$ be a
distribution on $[k]$.  The notation $W \sim w$ indicates that the
random variable $W$ is chosen according to $w$, i.e., $\Pr[W = i] =
w_i$.

\subsection{Process definition}

LAMP is parameterized by a matrix $P$ and a distribution $w$ on $[k]$.
The process evolves similarly to a Markov process, except that at time
$t$, the next state is chosen by drawing an integer $i$ according to
$w$, and picking an appropriate transition from the state $i$
timesteps ago (or from the start state if $t-i<1$).  Formally, we
define it as follows.

\begin{defn}[LAMP]
  Given a stochastic matrix $P$ and a distribution $w$ on $[k]$, the
  \emph{$k$th-order linearly additive Markov process} $\LAMP_k(w, P)$
  evolves according to the following rule:
\begin{equation}
\label{eqn:lamp}
\Pr[ X_t = x_t \mid x_0, \ldots, x_{t-1} ] = \sum_{i=1}^{k}
w_{i} \cdot P(x_{\max\{0,t-i\}},x_{t}).
\end{equation}
\end{defn}
When $k$ is implicit, we simply denote the process as $\LAMP(w, P)$.
We assume that the process starts in some state $\pi_0$. (As discussed
later, the particular choice of starting state is unimportant.)


\subsection{Key features}

(i) LAMP is a superset of first-order Markov processes; just set $w_1
= 1$.  We will show (Lemma~\ref{lem:lampvsfomc}) below that this is a
strict containment, i.e., there are processes expressible with LAMP
but provably not so with a first-order Markov process (despite the
comparable number of parameters).  We also explore other expressivity
relations between Markov processes and LAMP (Section~\ref{sec:express}).

(ii) By controlling the distribution $w$, one can encode highly
contextual transition behavior and long-range dependencies.  The
definition of $w$ determines how fast past history is ``forgotten'',
and allows a smooth transition from treating the recent past as a
sequence, to treating the more distant past as a set, simply by
flattening $w$.

(iii) While LAMP is a generative sequence model that allows long-range
dependencies, the model complexity is only $\nnz(P) + k$. In contrast,
the complexity of a $k$th-order Markov process is $\nnz(P_k)$, which
can be exponential in $k$.  In natural extensions of LAMP, the
distribution $w$ may be restricted to a family of distributions
specified only by a constant number of shape parameters (e.g., power
law, double Pareto, or log normal distributions).

(iv) An alternative characterization of LAMP provides a tractable and
interpretable way to understand its behavior
(Section~\ref{sec:analysis}).

(v) The LAMP model is easy to learn from data
(Section~\ref{sec:learn}).  While LAMP is by construction intended to
be a learned model with a tractable algorithm for the learning, the
mathematical structure of the model belongs to a much larger and
generally non-learnable Additive Markov processes
(Section~\ref{sec:disc}).


\subsection{Expressivity}
\label{sec:express}

A natural starting point to understand LAMP is to compare it in terms
of \textit{expressivity} and \textit{approximability} with Markov
processes. We begin by asking how closely LAMP, which employs a
stochastic matrix $P$, may be approximated by a first-order Markov
process.
\begin{lemma}
\label{lem:lampvsfomc}
There is a second-order LAMP that cannot be expressed by any
first-order Markov process.
\end{lemma}

\begin{proof}
  Define $\LAMP(w, P)$ as follows.  The transition matrix $P$ consists
  of a cycle on $\CX = [n]$, with $P(i, i+1) = 1$ for $i \neq 1$, and
  $P(1,1) = \epsilon, P(1,2) = 1 - \epsilon$ for some small $\epsilon
  > 0$ (this is to ensure mixing).  Let $k = 2$ and $w = (1/2, 1/2)$.
  All the additions are modulo $n$.  

  By the definition of $w$ and (\ref{eqn:lamp}), we can see that a
  sequence generated by $\LAMP(w, P)$ can contain the pattern $\ldots,
  i - 1, i, i, \ldots$ but can never contain the pattern $\ldots, i,
  i, i, \ldots$.  I.e., $\LAMP(w, P)$ can produce two repeats of a
  state but not three.  However, if we try to represent this process
  with a first-order Markov process $P_1$, then consider $P_1(i, i)$.
  If $P_1(i, i) = 0$, then the first-order process can never generate
  the first pattern and if $P_1(i, i) > 0$, then the first-order
  process generates the second pattern with positive probability.
  Neither of them is correct compared to $\LAMP(w, P)$.
\end{proof}
This example can be easily extended to show that there is a
$k$th-order LAMP that cannot be well approximated by any
$(k-1)$st-order Markov process. In summary, though the additional
parameters of a $\LAMP(w,P)$ are few, they are sufficient to give LAMP
more degrees of expressivity.

Of course, if the order of the Markov process is high enough, then it
strictly contains all LAMPs of a particular order. The following
statements make this precise.

%
\begin{lemma}\label{lem:lamp-in-markov}
$k$th-order LAMPs can be expressed by $k$th-order Markov processes.
\end{lemma}
\begin{proof}
  From (\ref{eqn:lamp}), that the transitions of $\LAMP_k(w, P)$
  depend only on $k$ tuples, since $w$ has a support of size $k$.
  Therefore, one can trivially use a $k$th-order Markov process to
  (wastefully) represent the transition probabilities.
\end{proof}
%
%
\begin{lemma}
  For $k\geq2$, there is a $k$th-order Markov process that is
  not expressible by a $k$th-order LAMP.
\end{lemma}
\begin{proof}
For sake of simplicity, we present the proof for $k = 2$; the general
case can be proved using similar ideas.

We consider a second-order Markov process on two states, $\CX = \{x,
y\}$.  We explicitly construct the following $P_2$ that is not
expressible as a second-order LAMP.  Let $\beta \neq \gamma$ and
$\alpha = 1 - \delta$.

\centerline{
\begin{tabular}{c|cc}
\hline
  $P_2$   & $x$ & $y$ \\
\hline
$\tuple{x, x}$ & $\alpha$ & $1 - \alpha$ \\
$\tuple{x, y}$ & $\beta$ & $1 - \beta$ \\
$\tuple{y, x}$ & $\gamma$ & $1 - \gamma$ \\
$\tuple{y, y}$ & $\delta$ & $1 - \delta$ \\
\hline
\end{tabular}
}

Suppose $P_2$ is expressible by $\LAMP_2(w, P)$ for some $w$ on $[2]$
and some stochastic matrix $P$.  We first claim that $P(x, x) =
\alpha$ and $P(y, y) = \delta$, i.e., the diagonals are fixed.  This
follows from the definition of $P_2$ and (\ref{eqn:lamp}) since, for
example,
$$
\alpha = P_2(\tuple{x, x}, x) = \Pr[x \mid \tuple{x, x}] = \sum_{i =
  1}^2  w_i P(x, x) = P(x, x).
$$
For the non-diagonals, we can compute
\mymath{
\beta = P_2(\tuple{x, y}, x) 
= w_1 \alpha + (1 - w_1) (1 - \delta),}
\mymath{
\gamma = P_2(\tuple{y, x}, x) 
= w_1 (1 - \delta) + (1 - w_1) \alpha.
}
It is easy to check that the above system is inconsistent with the
imposed conditions $\beta \neq \gamma$ and $\alpha = 1 - \delta$.
\end{proof}

Note that this result that $\LAMP_k \subsetneq \MC_k$ is also
intuitive, as a $k$th-order Markov Process has $O(n^k)$ parameters,
compared to a $k$th-order LAMP, which only has $n^2 + k$
parameters. The surprising aspect is that incorporating a little more
trail history with just a few more parameters leads to
inapproximability by lower order Markov processes, i.e., $\LAMP_k
\not\subseteq \MC_{k-1}$.


\section{Analysis of LAMP}
\label{sec:analysis}

Despite the differences in expressivity of LAMP and Markov processes,
we show that under fairly general conditions they share the same
limiting distribution over states, although the limiting distribution
on paths of length more than 1 may be very different.

As LAMP's evolution depends on history, we cannot simply characterize
its limiting distribution as a fixed point distribution on the graph.
Instead, we say that $\LAMP(w, P)$ has a \emph{limiting distribution}
$\pi$ if $\pi$ is a probability distribution over all states and
$\pi(x)$ is the probability of finding the LAMP at state $x \in \CX$
in the limit.  Next, we study the mixing time of $\LAMP(w, P)$ and
relate it to the mixing time of $P$ and the properties of $w$.  Mixing
time for LAMP is defined analogously to Markov processes; we denote
this by $\tmix(\LAMP(w, P), \delta)$. This study also leads us to a
connection between the evolution of $\LAMP(w, P)$ and the theory of
stochastic processes, particularly martingales and renewal
theory. This enables us to invoke results from these areas, and
precisely describe the long-term behavior of LAMP.  Besides being of
theoretical interest, knowing the mixing time of LAMP is important to
make conclusions about the validity of the learned model on real data.
In particular, we study the conditions under which mixing time of LAMP
can significantly depart from that of a Markov process.

\subsection{Exponent process}

To shed light on $\LAMP(w, P)$, we present a re-characterization of
the process in terms of certain exponents and mixtures of powers of
$P$.  Suppose $\pi_0$ is an initial distribution on $\CX$ and suppose
we run $\LAMP_2(w, P)$.  After the first step, the state distribution
is given by $\pi_0 P$.  After the second step, with probability $w_1$
we have the distribution $\pi_0 P^2$ and with probability $1 - w_1$ we
moved forward from $\pi_0$ again, and have the distribution $\pi_0 P$.
After the third step, with probability $w_1 w_2$ we have the
distribution $\pi_0 P^3$, with probability $(1 - w_1) w_2 + w_1 (1 -
w_2)$ we have the distribution $\pi_0 P^2$, and so on.  In this
interpretation, the distribution of a Markov process after $t$ steps
will be as $\pi_0P^t$, while the distribution of LAMP will be the
expectation of a random variable $\pi_0P^{e_t}$, where $e_t$ depends
on the random choices made from the history distribution, and may grow
and shrink over time.  We focus on the evolution of this exponent.

More formally, let $e_t$ be the random variable denoting the exponent
at time $t$.  The next state of the process is determined by choosing
an earlier state $t-i$ according to $w$, itself distributed as
$\pi_0P^{e_{t-i}}$, and then advancing one step from this state.  Once
the choice has been made to copy from $i$ states earlier, The new
exponent will therefore be $e_{t} = 1 + e_{t-i}$.  From this
intuition, we define the following stochastic process, which we use to
analyze size of the exponents.

\begin{defn}[Exponent process]
\label{exp-process}
Let $W_1, W_2, \ldots$ be i.i.d. random variables where $W_t \sim w$.
For $t \in \mathbb{Z}$, the \emph{exponent process} is the sequence of
random variables $e_t$, with $e_t = 0$ for $t \leq 0$, $e_1 = 1$, and
for $t > 1$:
\mymath{
   \Pr[e_t = e_{t -i} + 1] = \Pr[W_t = i].
}
\end{defn}

Note that for a first-order Markov process, the exponent process is
deterministic and is simply $...,0, 1, 2, \ldots$.  For LAMP, the
exponent process is more complex.  In the following sections, we
explore this process further, and discover not only concrete
statements about the limiting behavior and convergence to
equilibrium, but also connections to the theory of renewal
processes. As $e_t$ is pre-defined for $t \leq 0$, from henceforth we
assume $t \geq 0$.

We begin with a simple observation on the growth rate. 
\begin{lemma}
\label{lem:naive-exp-growth}
$e_t\geq\lfloor t/k \rfloor$.
\end{lemma}
\begin{proof}
  We can break the exponent process into consecutive batches of length
  $k$, where batch $B_i =\{e_{ik+1},\ldots,e_{(i+1)k}\}$.  Then
  $\min\{B_i\} \geq 1 + \min\{B_{i-1}\}$, since $e_t = 1 + e_j$ for
  some $j$ in $\{ t -k, \ldots, t -1 \}$.  Thus, the minimum exponent
  must grow by 1 every $k$ steps.
\end{proof}
This immediately gives the limiting distribution
of LAMP:
\begin{thm}
\label{equib-thm}
$\LAMP(w, P)$ has a limiting distribution $\pi$ if and only if $P$ is
ergodic.  Furthermore, $\pi$ is also the stationary distribution for
the first-order Markov process with transition matrix $P$.
\end{thm}
\begin{proof}
  The distribution of LAMP is given by the product of the start state
  with a mixture of powers of the transition matrix $P$ of the
  underlying Markov process.  Lemma~\ref{lem:naive-exp-growth} gives a
  lower bound on every exponent in the mixture, and this lower bound
  increases without bound.
\end{proof}
Note that the proof above only addresses the finite $k$ case. A
similar result can be stated, with probability almost surely, for the
infinite $k$ case, using results similar to Section 4.2. Theorem
\ref{equib-thm} also has a direct proof, which can be used to show a
similar limiting distribution result for a generalization of LAMP 
(details omitted).

%
%
Next, we study the mixing time of LAMP in more detail.
Lemma~\ref{lem:naive-exp-growth} gives us a bound on the mixing time,
but this bound characterizes the worst case. We now seek a stronger
bound based on the average case evolution of the exponent process.

\subsection{Analysis with renewal theory}

\begin{figure}
\centering
\includegraphics[width=1.0\columnwidth]{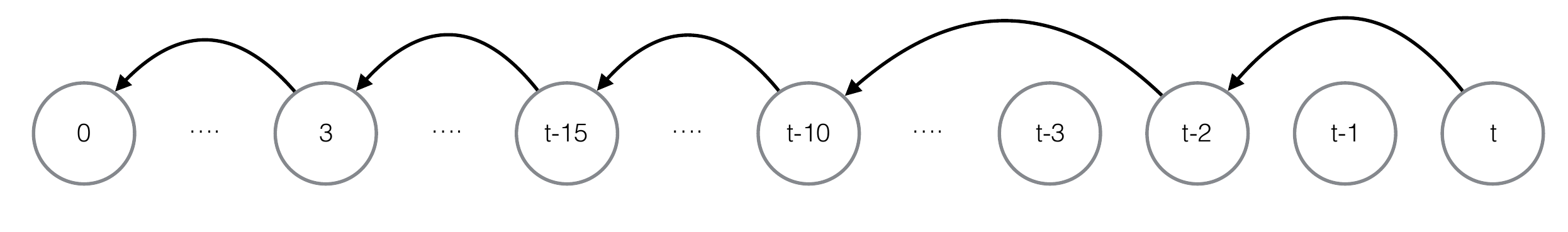}
\caption{Recovering backward chain from time $t$:  $t_0=t, t_1 = t-2,
  t_2 = t-10,\ldots, t_5=0$.  Here, $H(t) = 5$.}
\label{fig:backward-arrows}
\end{figure}

Consider the exponent process at a particular time $t$, with exponent
$e_t$.  We may trace the origins of this exponent backwards through
the evolution of the process.  At time $t$, the process chose a point
$t-i$ from $i$ steps in the past, and added 1 to the exponent at that
time.  Similarly, at time $t-i$, the process chose another point even
farther in the past from which to step forward.  This backward
chaining process is depicted in Figure~\ref{fig:backward-arrows}.  The
number of hops traversed backward at each step is selected i.i.d.\
from $w$.  For any point $t$, we may therefore define a sequence of
decreasing positions $t=t_0 > t_1 > \cdots > t_{H(t)}$, with
\mymath{
e_{t_i} = e_{t_{i + 1}} + 1,
}
and $H(t)$ a \textit{stopping time}, with $t_{H(t)}$ the first term
$\leq 0$. The total number of hops in this evolution is exactly $H(t)
= e_t$, as each step increases the exponent by 1.

We now recall the definition of a renewal process, and show that the
process defined above is a renewal process:

\begin{defn}[Renewal process]
  Let $S_1, S_2, \ldots$ be a set of positive i.i.d.\ random
  variables, and for $n \in \mathbb{Z}^+$ define the $n$th jump time $J_n =
  \sum_{i=1}^n S_i$.  Then the random variable $(X_t)_{t\geq 0}$
  defined by $X_t = \sup\{n ~\mid~ J_n \leq t\}$ is called a \emph{renewal
    process}.
\end{defn}

The connection to LAMP is made explicit as follows:

(i) The gaps $t_i - t_{i+1}$, each drawn i.i.d.\ from $w$,
  correspond to the holding time variables $S_i$.

(ii) $H(t)$ correspond to the random variables $X_t$ forming the
  renewal process.






  Based on this connection, we can employ the strong law of large
  numbers for renewal processes~\cite{resnick} to show the following
  theorem:
\begin{thm}
\[ \lim_{t \rightarrow \infty} \frac{H(t)}{t}  =
\frac{1}{E[w]}, \hspace{5mm} \mbox{almost surely}. \]
\end{thm}
Note that this bound is much stronger than the bound of
Lemma~\ref{lem:naive-exp-growth}.\footnote{A weaker statement of
  convergence in mean may be shown by using Wald`s
  identity~\cite{wald} combined with $H(t)$ being a stopping time to
  condition on the sum of i.i.d. random variables.  However, the law
  of large numbers allows us to make a statement regarding the random
  variable $H(t)$ rather than simply its expectation.}  Based on this
bound, we see that the mixing time of LAMP is related to the mixing
time of the underlying Markov process by a multiplicative factor of
$E[w]$ as long as the renewal process has attained its limiting
behavior.

One can then refer to the well-known central limit
Theorem~\cite{resnick} for renewal processes to get concentration
bounds.
\begin{thm}
If $W$ has finite mean and variance $\mu$ and $\sigma$, then
\mymath{
\lim_{t \rightarrow \infty} \Pr \left[ \frac{H(t) - t/\mu }{\sigma \mu^{-3/2} \sqrt{t}} < \alpha \right] = \phi(\alpha),}
where $\phi$ is the cdf of the standard Gaussian distribution.
\end{thm}
However, with slightly stronger assumptions on the moments of $W$, we
can also obtain growth rates for finite $t$.

\subsection{Concrete growth rates and mixing time}
\label{concrete-growth-rates}

To address our original question on mixing times, we derive the
following concrete growth rate via Bernstein's
inequality~\cite{DP}. 
\begin{thm}
\label{thm-growth-rate}
For $k$ finite, for all $t \geq T$, 
\mymath{
\Pr \left[ e_t  \geq \frac{t}{(1 + \epsilon)E[w]} \right] \geq 1 -
\frac{ e^{-C(\epsilon) \cdot T}} {1 - e^{-C(\epsilon)}},}
where $C(\epsilon)$ is a constant depending on $\epsilon, k$, and some
of the moments of $w$. 
\end{thm}
this version.)  This immediately bounds the mixing time.
\begin{thm}
For $w$ with finite support $[k]$, we have
\mymath{
\tmix(\LAMP(w, P), \delta) 
  \leq \max\left\{ T, (1 + \epsilon)E[w] \cdot \tmix(P, \delta) \right\},}
with probability at least
$1 - \frac{e^{-C(\epsilon) \cdot T}}{1 - e^{-C(\epsilon)}}$.
\end{thm}
We can also establish similar results (formal statement and proofs
omitted in this version) for infinite $k$ using the following:
\begin{thm} 
Let $E[|w|^4] \leq \infty$.  Then, for all $t \geq T$, 
\mymath{
\Pr \left[ e_t \geq \frac{t}{(1 + \epsilon)E[w]} \right] \geq 1 - O\left( \frac{1}{T} \right).
}
\end{thm}
With these results, we conclusively establish long-term behavior of
$\LAMP(w, P)$, and next look at learning the model.

\section{Learning the parameters}
\label{sec:learn}

We now show how to learn the parameters of LAMP, given a sequence of
observations. The input to the learning problem is a sequence $x_0,
\ldots, x_t$ of states, on a state space $\CX$. The goal is to learn
the transition matrix $P$ and the distribution $w$ that maximizes the
likelihood of the observed data.

Given the sequence of states, by the definition of LAMP, we have
\mymath{
\Pr[ x_{k+1} \mid x_1, \ldots, x_{k} ]
 = \sum_{i=1}^{k}w_{i} \cdot P(x_{k+1-i}, x_{k+1}).
}
In particular, the likelihood function decomposes into a product of
such terms and so the log likelihood is
\mymath{
L(x_0,...,x_t) 
= \sum_{j=0}^t \log \left(\sum_{i=1}^{k}w_{i} \cdot P(x_{j-i},x_j) \right).
}
From here onwards we write $x_{0}$ in place of $x_{\max\{0,l\}}$ to simplify the notation.

To compute the gradient with respect to the entries of the transition
matrix $P$, we compute the gradient with respect to each term in the
above expression and put it together. Let $I(\cdot)$ denote the binary
indicator function. We first have
\mymath{
\frac{\partial L}{\partial P(x, y)}
   = \sum_{x_j = y} \displaystyle{ \frac{\sum_{i=1}^{k} I(x_{j-i} = x)
       \cdot w_i}{\sum_{i'=1}^{k}w_{i'}P(x_{j-i'},x_j)}}.
}
For the entries of $w$, we have
\mymath{
\frac{\partial L}{\partial w_i} 
= \sum_{j=0}^t \frac{P(x_{j-i}, x_j)} {\sum_{i'=1}^{k} w_{i'}
  P(x_{j-i'}, x_j)}.
}

It is easy to show that the log likelihood is individually concave in
$P(\cdot, \cdot)$ and $w$ but not jointly concave (proofs omitted). We
run an alternating minimization to optimize for $P$ and $w$ holding
the other parameters fixed. Recall that $w$ and every row of $P$ is
non-negative and sums to $1$. Because of the latter constraint,
generic unconstrained or box-constrained optimization algorithms such
as L-BFGS(-B) cannot be applied to our problem directly.

Let us consider optimizing $w$ while holding $P$ fixed first. By the
Karush--Kuhn--Tucker (KKT) optimality conditions we have that $w^{*}$
is optimum if and only if there exist $\lambda > 0$,
Lagrange-multiplier, such that $\frac{\partial L}{\partial w_i}(w^*) =
\lambda$ if $w_i^{*} > 0$ and $\frac{\partial L}{\partial w_i}(w^*)
\le \lambda$ if $w_i^{*} = 0$. Generally, similar KKT conditions can
be solved with complicated sequential quadratic programming or with
interior-point methods~\cite{nocedal2006numerical}.  Instead, we
recognize that our KKT condition is a non-trivial extension of the
\emph{water-filling} problem~\cite[Example 5.2]{boyd2004convex}, that
arises in information theory in allocating power to a set of
communication channels. Starting from $\lambda=\infty$, attained by
setting all $w_i^*=0$, we swipe with $\lambda$ towards $0$ until we
find a $\lambda$ value satisfying the KKT conditions with
corresponding $w^*$ that sums to one. To compute changes in
$\frac{\partial L}{\partial w_i}$ we rely on the Hessian of $L$, i.e.,
we apply Newton's method. Furthermore, we pretend that $H$ is
diagonal.  While this assumption is clearly false for $w$, the
approximation is good enough that the resulting method works well in
practice.

    Then given an initial guess, $w^{0}$, our goal is to find an
    adjustment $u$ such that $w^{0}+u$ is still feasible, i.e., $\sum_i
    u_i = 0$ and $0 \le w^{0}_i + u_i$ hold for all $i=1, \ldots, k$,
    and the
    KKT condition is satisfied by $w^{0}+u$. Such $u$ can be found
    with the aforementioned water-filling technique relying on the
    fact that each approximate $\frac{\partial L}{\partial w_i}$
    becomes a linear function. We iteratively apply these adjustments
    until the KKT condition is satisfied up to a small error. Since
    our Hessian approximation introduces errors, it can happen that an
    adjustment would decrease the log likelihood. To combat this
    issue, we apply the well-known \emph{trust region method}
    \cite{nocedal2006numerical} and search for an adjustment $u$ with
    $\max_i{|u_i|} \le r_t$, where $r_t$ is the size of the trust
    region in the $t$th adjustment step. Note that optimizing over
    $P$ decomposes over optimizing rows of $P(x, \cdot)$, where optimizing
    each row $P(x, \cdot)$ is analogous to optimizing $w$. Furthermore
    it is not hard to see that the Hessian is indeed diagonal in this case.
    Thus we optimize each row of $P(x, \cdot)$ while holding the rest
    fixed, i.e., we perform block coordinate descent.

    Lastly we remark that Duchi et al.'s projected gradient method
    \cite{duchi2008efficient} is also applicable to our problem. We
    found it slow to converge as all gradients are equal and positive
    in optimum typically.

\newcommand{\lastfm}{{\sc Lastfm}\xspace}
\newcommand{\brightkite}{{\sc BrightKite}\xspace}
\newcommand{\wiki}{{\sc WikiSpeedia}\xspace}
\newcommand{\shakespeare}{{\sc Shakespeare}\xspace}
\newcommand{\reuters}{{\sc Reuters}\xspace}
\newcommand{\wsj}{{\sc WSJ}\xspace}

\section{Experiments}
\label{sec:exp}

In this section we present our experimental results on LAMP.  The goal
of these experiments is to model real-world sequences with LAMP and
compare their performance with first- and higher-order Markov
processes.  For comparing the performance, we focus on standard
notions such as perplexity.\footnote{
Perplexity of model $q$ on sequence $x_0,\ldots,x_t$ is
the reciprocal of the geometric mean of the probabilities of observed transitions:
$2^{\frac{-1}{t}\sum_{i=1}^t\log_2q(x_i|x_{i-1},\ldots,x_{i-k})}$;
the lower the better.}
En route we also evaluate the performance
of the learning algorithm based on alternative minimization, focusing
on the number of iterations, convergence, etc.  First we describe the
datasets used for evaluation.

\subsection{Data}

We use the following datasets.  Each dataset is a sequence of items
over a particular domain.  Since we are interested in modeling the
transitions, the items will correspond to the states in LAMP (and in
the Markov processes we will compare against) and we will not use any
metadata about the items themselves.  For repeatability purposes, we
only focus on publicly available data.

\para{\lastfm}
This data is derived from the listening habits of users on the music
streaming service \url{last.fm}~\cite{celma2009music}.  In this
service, users can select individual songs or listen to stations based
on a genre or artist.  We will focus on sequences, where each sequence
corresponds to a user and each item in the sequence is the artist for
the song (we focus on artists instead of individual songs).  The data
consists of 992 sequences with 19M items.  The data is available at
\url{dtic.upf.edu/~ocelma/MusicRecommendationDataset/lastfm-1K.html}.

\para{\brightkite}
BrightKite is a defunct location-based social networking website
(\url{brightkite.com}) where users could publicly check-in to various
locations.  Each sequence corresponds to a user and each item in the
sequence is the check-in location (i.e., the geographical coordinates)
of the user.  The data consists of 50K sequences with 4.7M check-ins.
This dataset is available at
\url{snap.stanford.edu/data/loc-brightkite.html}.

\para{\wiki}
This dataset is a set of navigation paths through Wikipedia, collected
as part of a human computation game~\cite{wiki1,
  wiki2}. In the game users were asked to navigate from a starting
Wikipedia page to a particular target only by following document
links. The dataset consists of 50K (source, destination) paths,
which will form the sequences for our experiments; the items in a
sequence will correspond to the Wikipedia pages.  The dataset is
available at \url{snap.stanford.edu/data/wikispeedia.html}.

\para{\reuters}
As an illustration of the performance of LAMP on a text dataset, we
consider the Reuters-21578, Distribution 1.0 benchmark corpus
(available at \url{www.nltk.org/nltk_data/}) as a baseline. Here, each
newswire article is considered to be a single sequence and the items
are the words in the sequence.  The data consists of more than 1.2M
words.  

To mitigate data issues, we focus on non-consecutive reasonably
frequent visits.  Since consecutive repetitions for \lastfm might mean
songs from the same CD and for \brightkite might mean being in the
same place---yielding easy to learn, mundane first-order models with
dominant self-loops as best fit,---we collapse consecutive repeated
items into a single item.  We then replace items that appear fewer
than 10 times (in the original sequence) by a generic `rare' item;
this threshold is 50 for \lastfm.

\subsection{Baselines}

We compare LAMP to various baselines, defined as follows.

\para{Naive $N$-gram} An order-$k$ Naive $N$-gram behaves as $\MC_k$,
i.e., a $k$th-order Markov process, which employs counts of the
previous $k$ elements to predict the next element.  Since a
second-order Markov process in fact uses statistics about 2 past and 1
future element, order-2 corresponds to 3-grams, and generally,
order-$k$ corresponds to $k+1$-grams.  We retain the notion of order
so that order-1 Naive $N$-gram will be equivalent to order-1 LAMP.

\para{Kneyser--Ney $N$-gram} This baseline employs state-of-the-art
smoothing of $N$-grams using the Kneyser--Ney
algorithm~\cite{kneser-ney}.  It corresponds to a variable-order
Markov process, in that it employs as much as context as possible,
given the available data, falling back to lower-order statistics as
necessary.

\para{LSTM} This baseline trains an LSTM (Long Short Term Memory)
architecture on the datasets. The LSTM is a type of recurrent neural
network~\cite{1997lstm}, which has recently been very successful in
modeling sequential data.

\para{Weight-only LAMP} This baseline corresponds to running the LAMP
algorithm with the matrix $P$ fixed to the empirical transition
matrix, learning only the weights $w$.

\para{Initial weights} This baseline corresponds to LAMP with the
matrix $P$ fixed to the empirical transition matrix and the history
distribution defined as $w_i \propto 0.8^i$.  We always initialized
the alternating minimization to these values and hence this baseline.

\subsection{Results}

\begin{figure}[htb]
\centering
\begin{tabular}{cc}
\includegraphics[width=0.5\columnwidth]{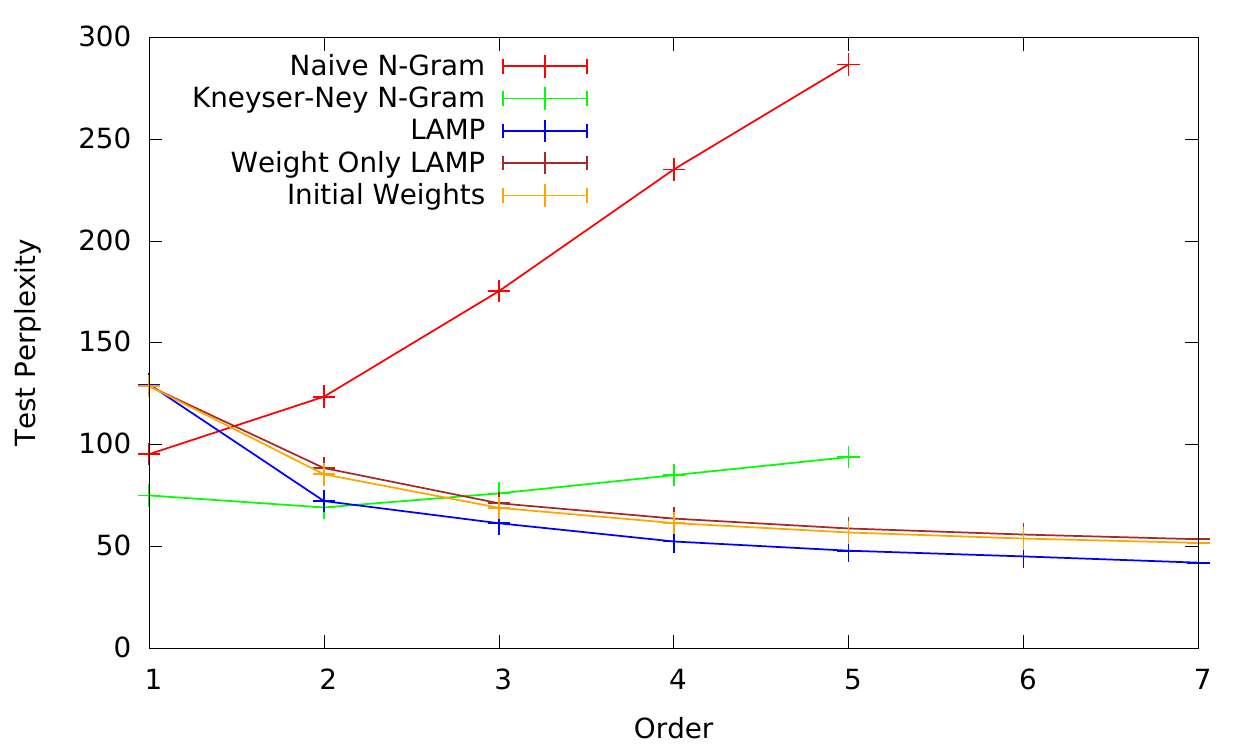}
&
\includegraphics[width=0.5\columnwidth]{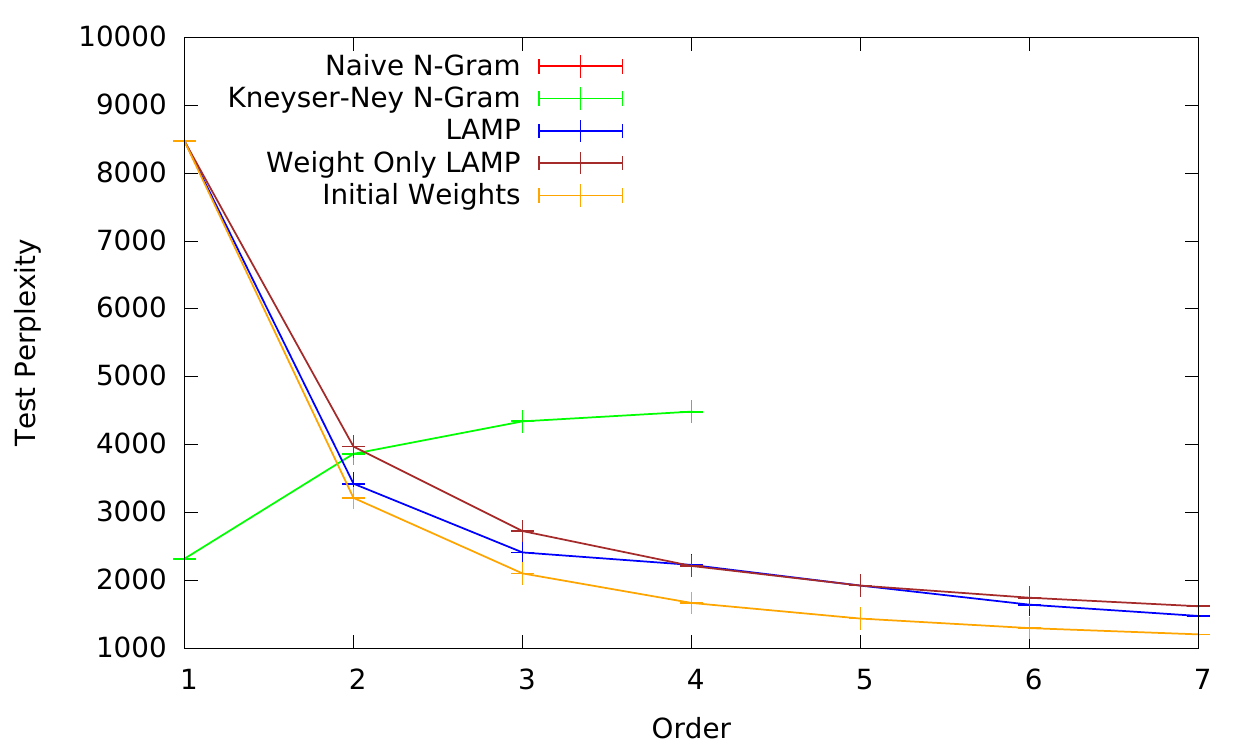} 
\\
\includegraphics[width=0.5\columnwidth]{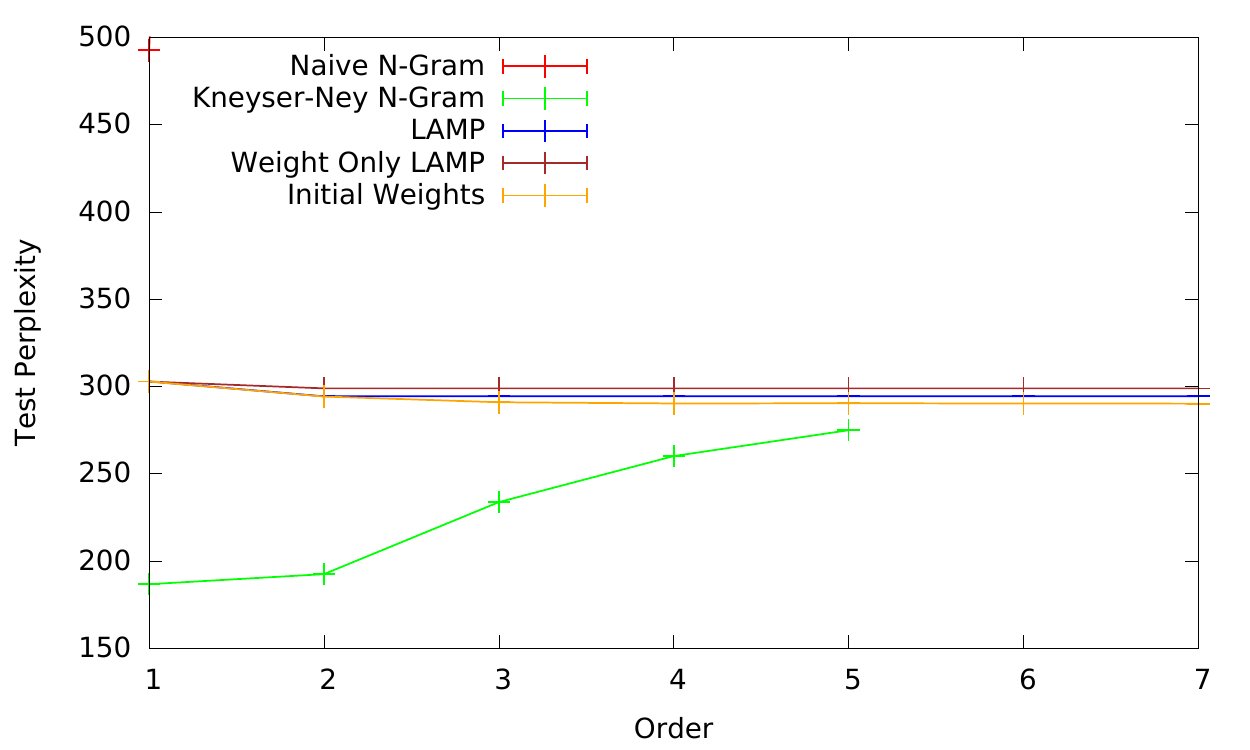}
&
\includegraphics[width=0.5\columnwidth]{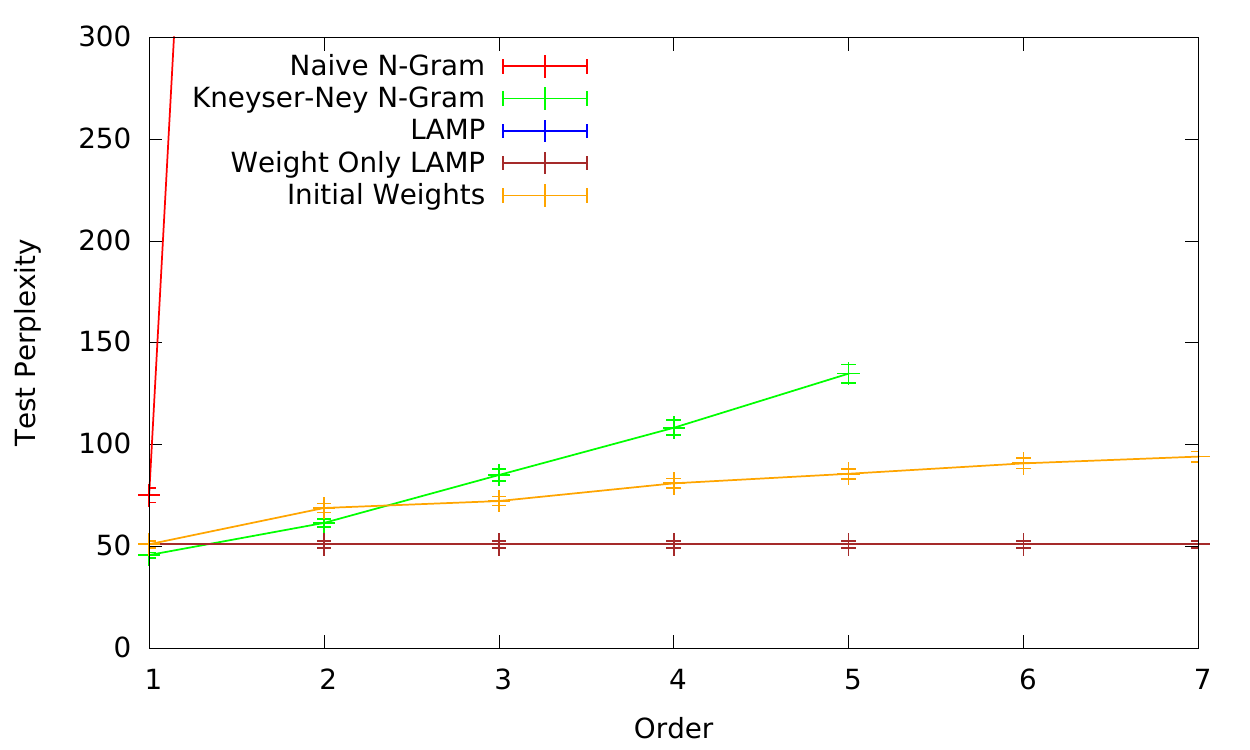}
\end{tabular}
\caption{Order vs test perplexity for LAMP, weight-only LAMP, MLE and
  Kneyser--Ney N-grams. Top left: \brightkite, top right: \lastfm, bottom
  left: \reuters, bottom right: \wiki.}
\label{fig:order-vs-perplexity}
\end{figure}

We compare LAMP to the $N$-gram variants and to the LSTM.

\subsubsection{LAMP and $N$-grams}

\para{Perplexity}
Figure~\ref{fig:order-vs-perplexity} gives our main results on the
perplexity of various order models, across our datasets.  The first
chart of the figure covers \brightkite\ data, and shows that all LAMP
variants outperform the Markov models on test perplexity, with an
order-7 LAMP showing around 19\% improvements over the simpler
variants at the same order.

For \lastfm, again LAMP variants outperform N-grams, producing
text perplexity around 50\% lower than smoothed N-grams, and far lower
than naive N-grams, which overfit badly.  In this case, among the LAMP
variants, the exponential weights improve over the optimized by about
18\% at order 7.  The optimized weights perform better than
exponential on training perplexity, so the algorithm could benefit
from some additional regularization.

For \reuters, LAMP models perform similarly to one another, and
the naive N-grams perform so poorly that they do not register on the
chart.  The smoothed N-grams outperform all LAMP variants.  However,
despite the smoothing, their performance on test worsens as the data
becomes too sparse; this is a regime where the smoothing is not
designed to operate.  For this data, LAMP sets $w_1=1$, behaving as a
first-order Markov process, and higher-order N-grams are able to
perform better, albeit with the usual trade-off of higher numbers of
parameters.\footnote{LAMP may of course employ a higher-order process
  as its underlying matrix, simply by exploding the state space.  It
  is unlikely in this case that it would do more than match the
  performance of smoothed N-grams, as more remote history without
  intervening context does not appear to work well for this dataset.}

Finally, for \wiki, naive and smoothed N-grams perform substantially
worse than LAMP variants, with LAMP and weight-only LAMP performing
best.

As described in Lemma~\ref{lem:lamp-in-markov}, $\MC_k$ is more
expressive than $\LAMP_k$, so one may wonder how it is possible that
order-$k$ LAMP may show better performance than a $k$th-order
Markov process.  In fact, the Markov process performs significantly
better in training perplexity, but despite our tuning of the
Kneser--Ney smoothing, LAMP generalizes better.

\begin{figure}[h]
\centering
\includegraphics[width=0.5\columnwidth]{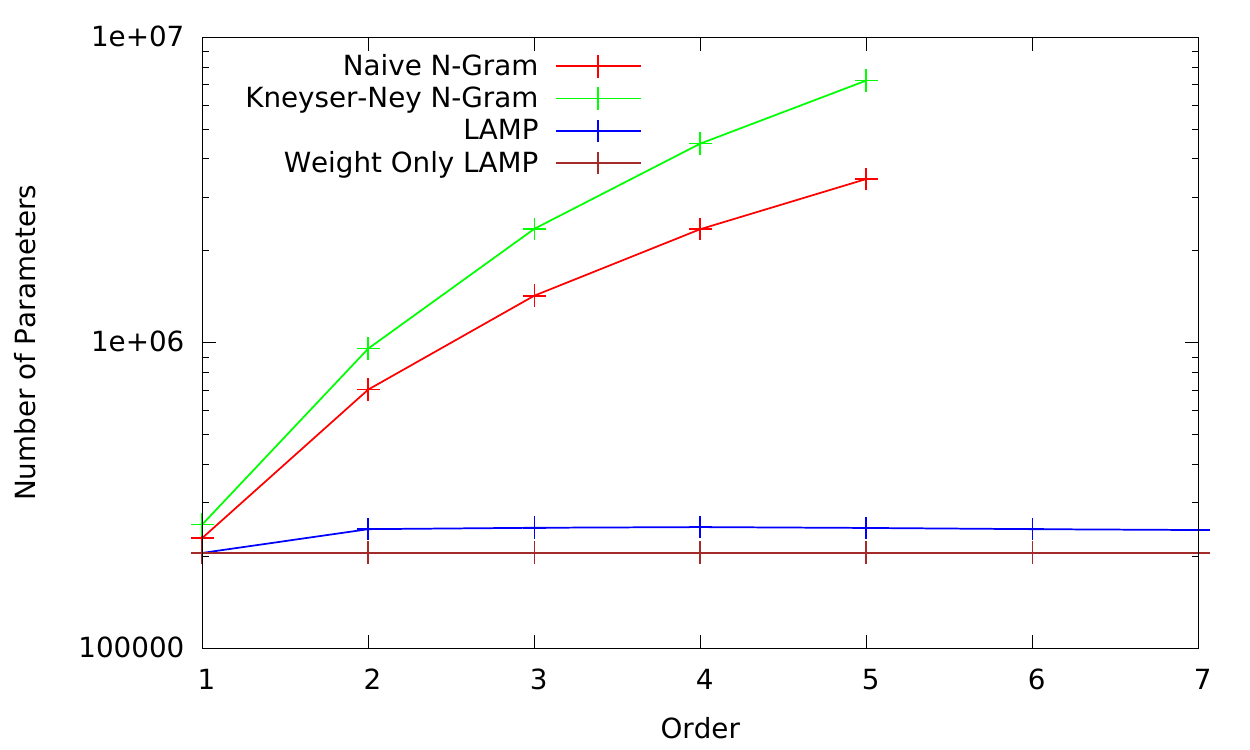}
\caption{Order vs number of parameters for \brightkite.  Other datasets
  are similar.}
\label{fig:order-vs-numparams}
\end{figure}

\para{Number of parameters}
Figure~\ref{fig:order-vs-numparams} shows how the actual number of
parameters changes with order for the various models.  For the LAMP
models, the number of parameters could theoretically grow as order
increases, because the optimizer will see possible transitions from
earlier states, and might choose to increase the density of the
learned matrix.  For example, on an expander graph, a sparse
transition matrix could even grow exponentially in density as a
function of order until the matrix becomes dense.  While this growth
is possible in theory, it does not occur in practice.  The
\brightkite\ dataset shows a small increase in parameters from first
to second order LAMP, and the remaining datasets show no significant
increase in parameters whatsoever.  The N-gram algorithms show in all
cases a dramatic increase in parameters, to the extent that we could
run these baselines only to order 5 (and order 3 for \lastfm).

\begin{figure}[ht]
\centering
\begin{tabular}{cc}
\includegraphics[width=0.5\columnwidth]{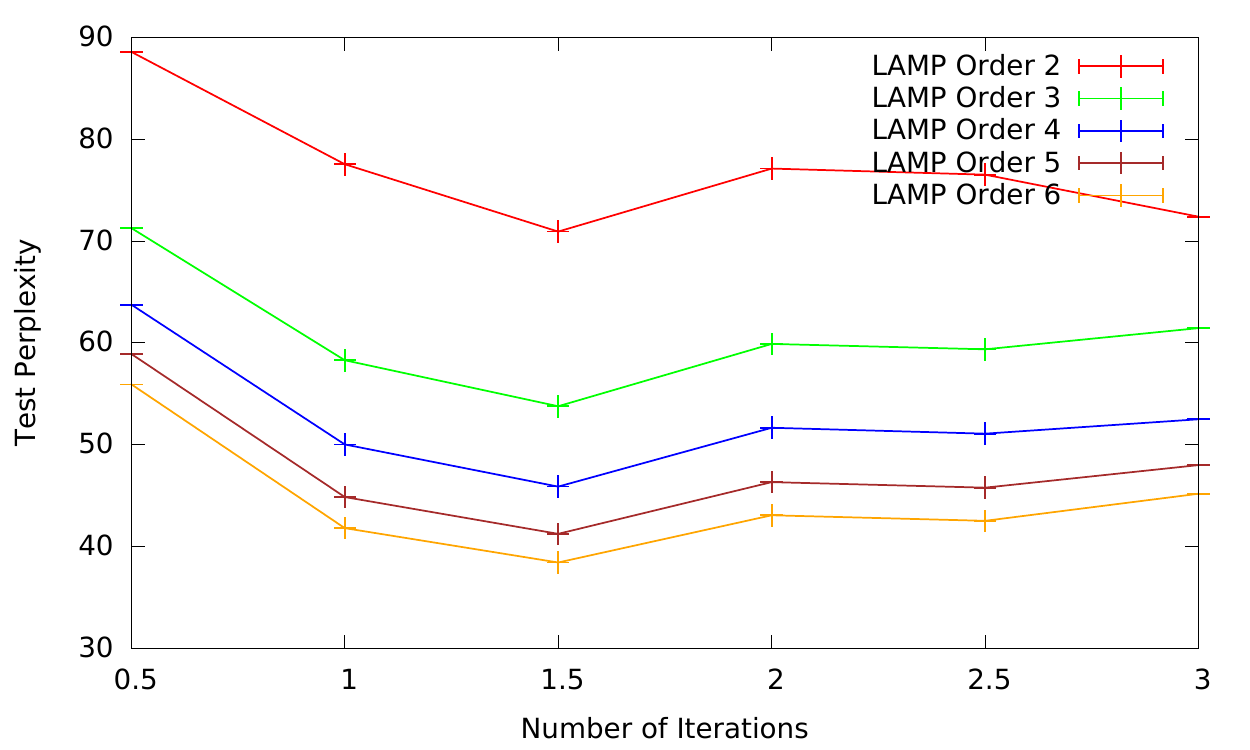} &
\includegraphics[width=0.5\columnwidth]{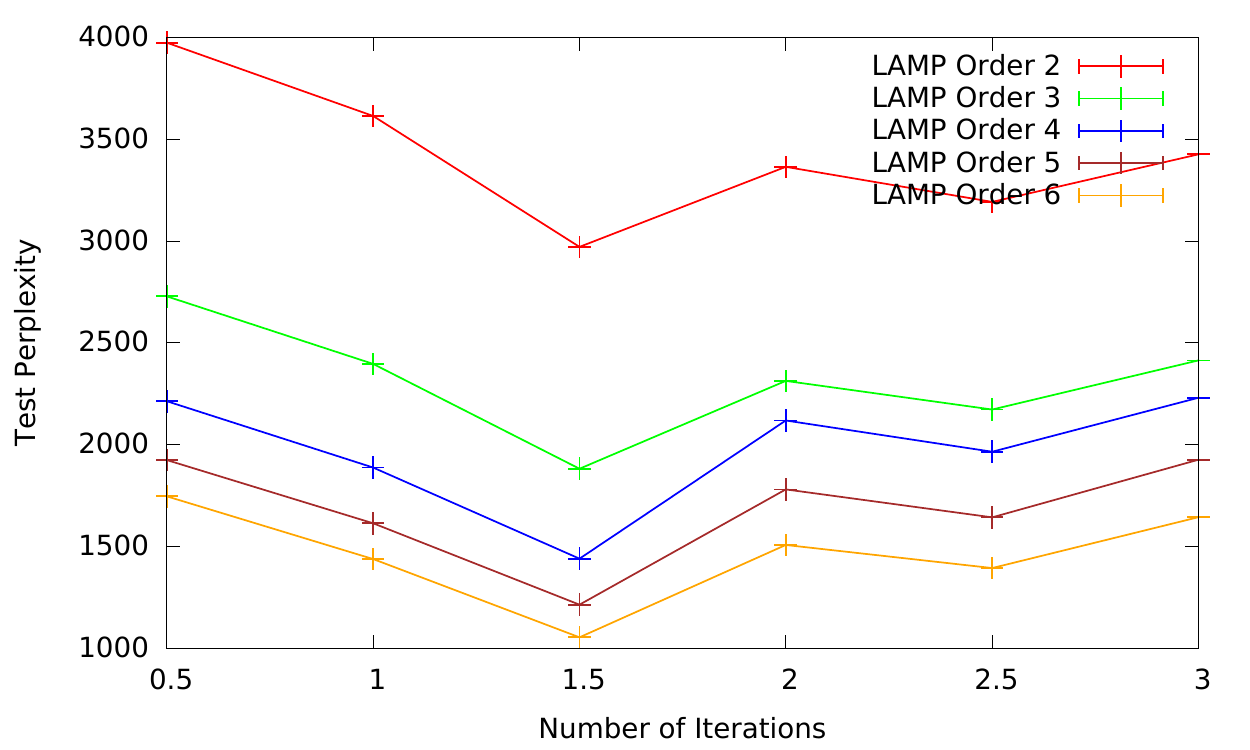} \\
\includegraphics[width=0.5\columnwidth]{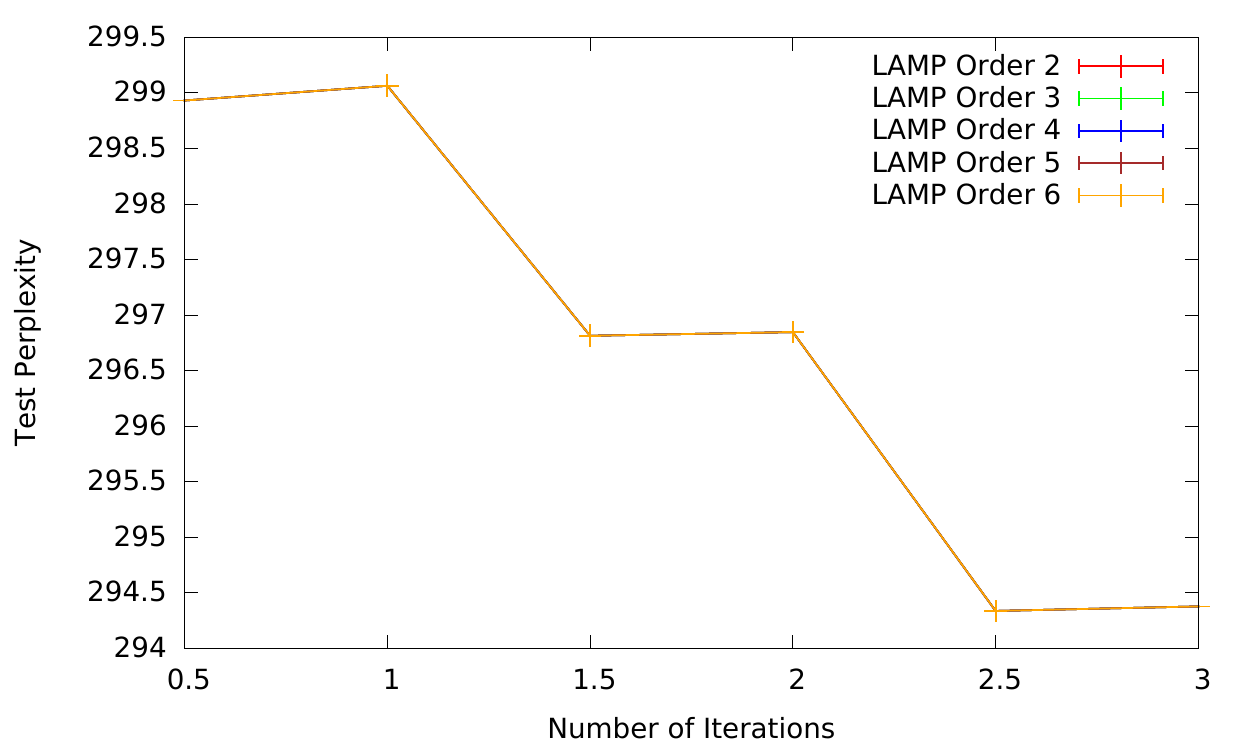} &
\includegraphics[width=0.5\columnwidth]{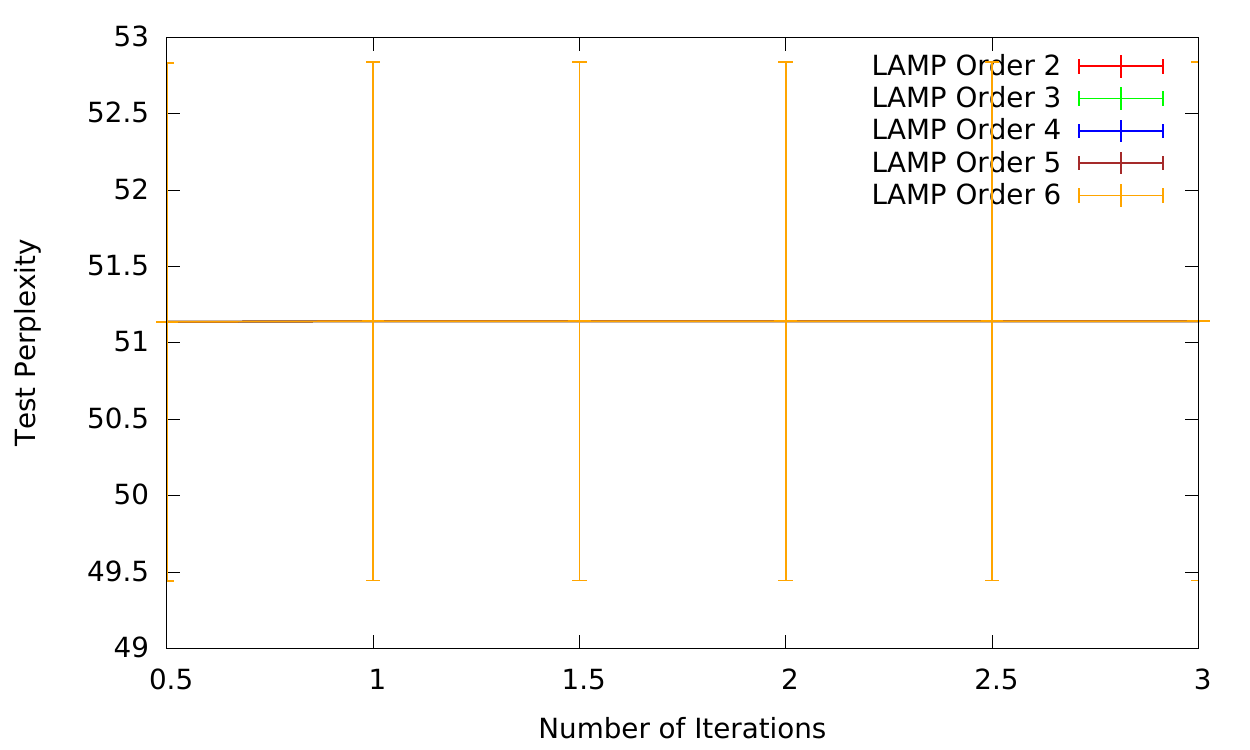}
\end{tabular}
\caption{Number of iterations in alternating minimization vs test
  perplexity for LAMPs of various order. $i - 0.5$ is the $i$th weigh
  optimization, $i$ is the $i$th transition matrix optimization. 
  Top left: \brightkite, top right: \lastfm, bottom left: \reuters, 
  bottom right: \wiki, error bars show standard deviation in $10$-fold cross
  validation; repeated for 2 datasets.
}
\label{fig:iter-vs-perplexity}
\end{figure}

\para{Number of rounds}
Figure~\ref{fig:iter-vs-perplexity} shows how the performance of LAMP
changes over the rounds of optimization.  For \brightkite\ and
\lastfm, the algorithm performs best at 1.5 rounds, meaning the
weights are optimized, then the matrix, and then the weights are
re-optimized for the new matrix.  For \reuters, small improvements
continue beyond this point, and performance for \wiki\ is flat
across iterations.  The appropriate stopping point can be determined
from a small holdout set, but from our experiments, performing 1.5
rounds of optimizations seems effective.

\begin{figure}[ht]
\centering
\begin{tabular}{cc}
\includegraphics[width=0.5\columnwidth]{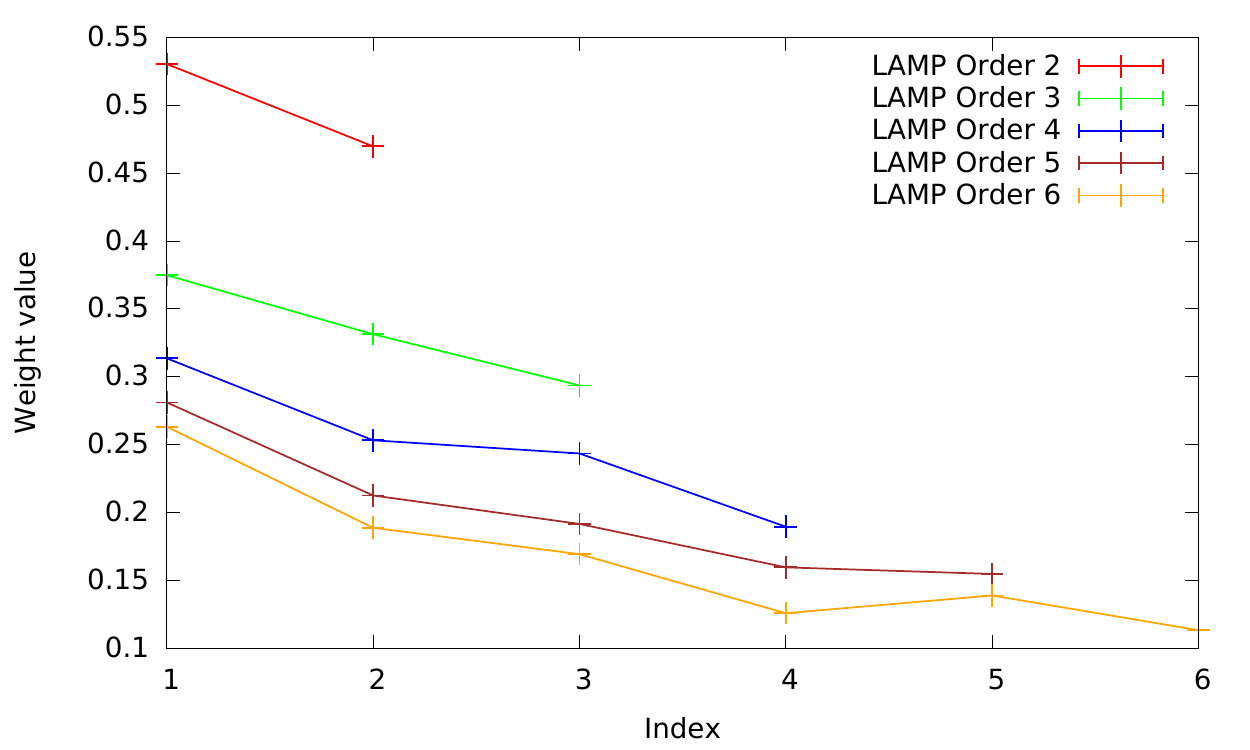} &
\includegraphics[width=0.5\columnwidth]{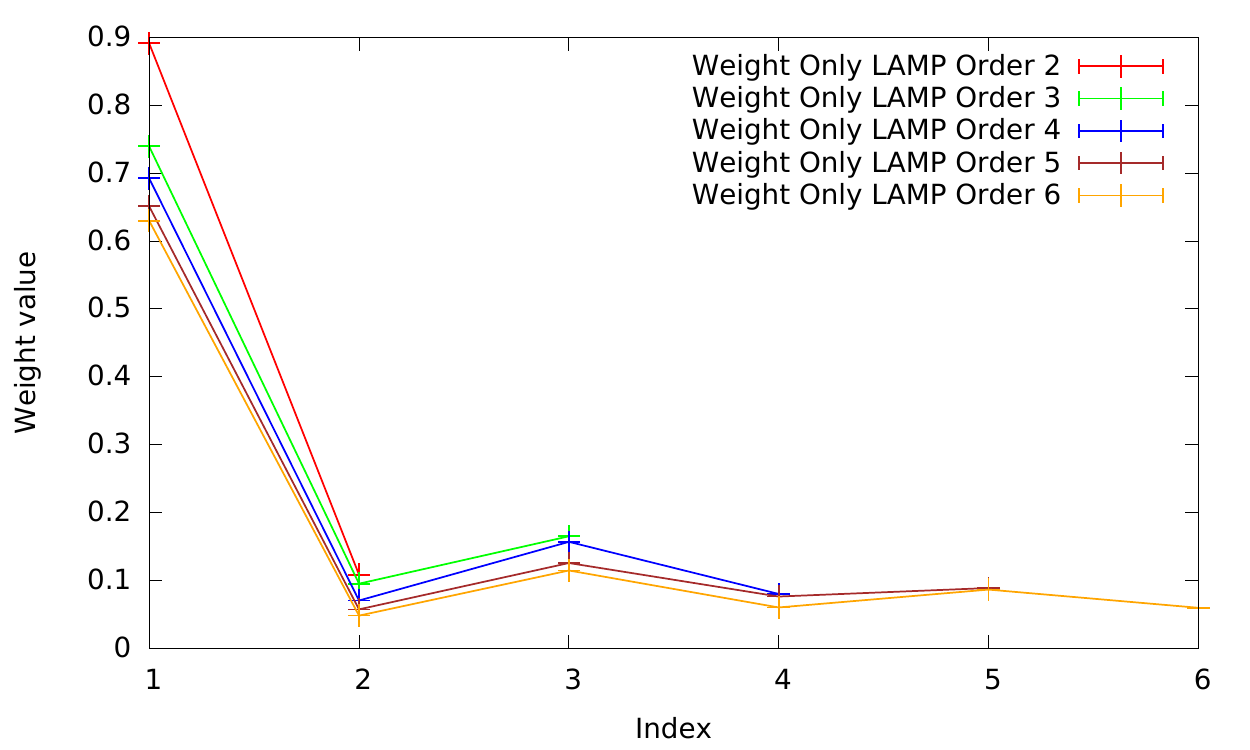} \\
\includegraphics[width=0.5\columnwidth]{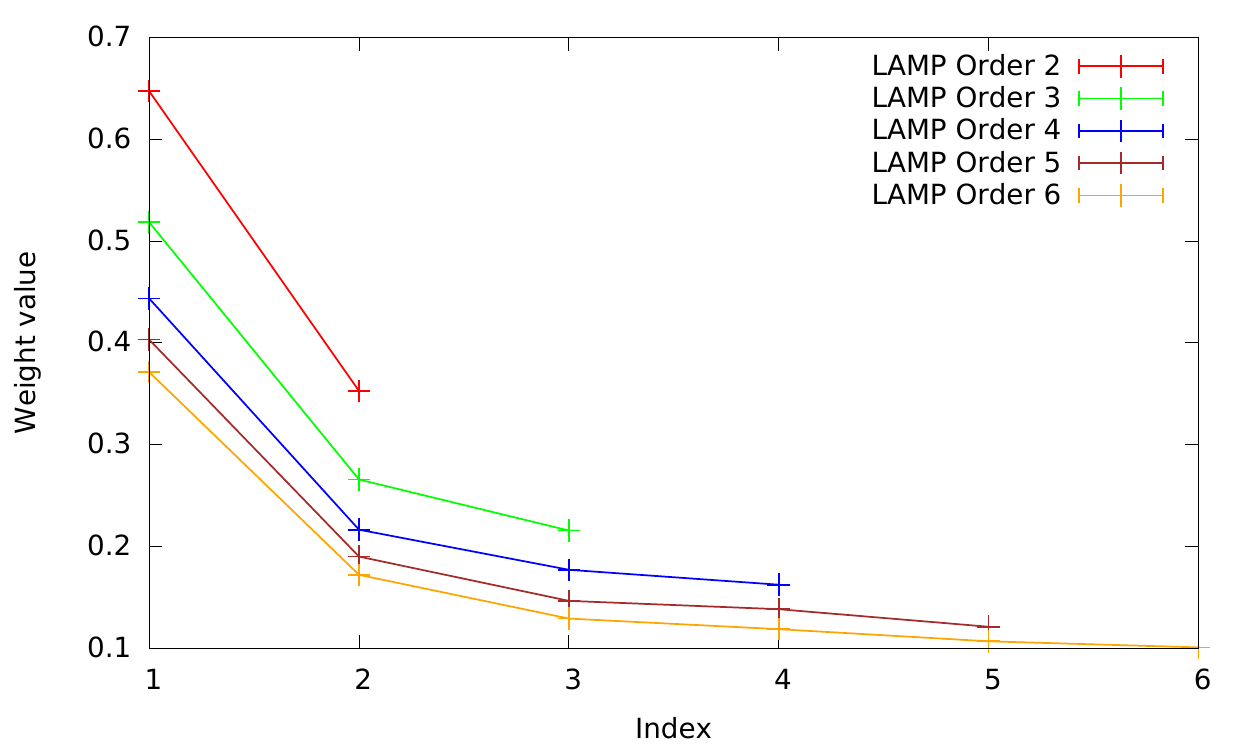} &
\includegraphics[width=0.5\columnwidth]{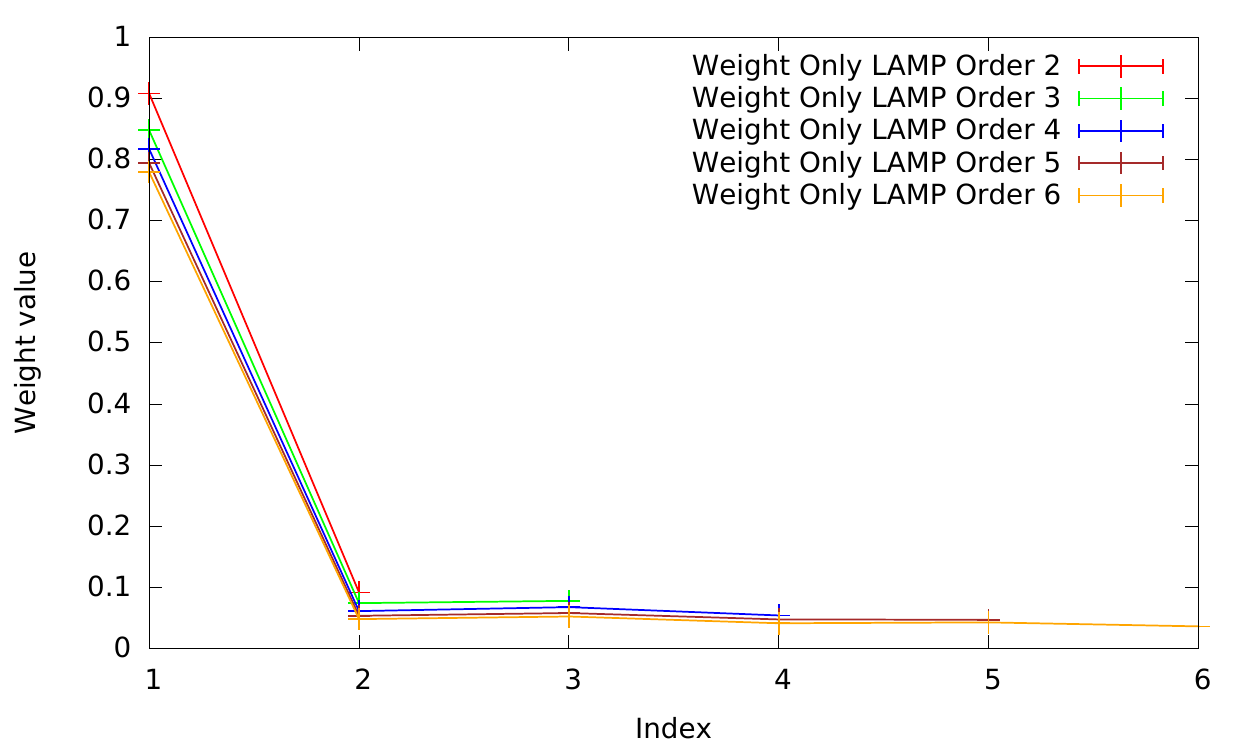}
\end{tabular}
\caption{Estimated weight values for LAMPs of various order (left) and
  weight-only LAMPs (right), Top left: \brightkite, top right: \lastfm,
  bottom left: \reuters, and bottom right: \wiki set $w_1=1$.}
\label{fig:weight-values}
\end{figure}

\para{Learned weights}
We now turn to the weight values learned by the LAMP model; see
Figure~\ref{fig:weight-values}.  \brightkite\ and \lastfm\ show a
similar pattern: when the matrix is fixed to the empirical observed
transitions (right column), the weights decay rapidly, putting most of
their focus on the most recent state.  This behavior is expected, as
the matrix in this case has been selected to favor the most recent
state.  However, when LAMP is given control of the matrix, the
optimizer selects a matrix for which a flatter weight distribution
improves likelihood, making better use of history.

We also observe that the relative values of earlier weights in
lower-order versus higher-order LAMP are similar.  As order increases,
all the weights are dropped slightly to allow some probability mass to
be moved to higher weights corresponding to influence from more
distant information.

Both \reuters\ and \wiki\ assign $w_1=1$, indicating that
LAMP's history is not beneficial in modeling these datasets.

\subsubsection{LAMP and LSTM}

To this point, we have considered LAMP as an alternative sequence
modeling technique to Markov processes, appropriate for use in similar
situations.  In this section, we consider the performance of LAMP
relative to LSTMs, which lie on the alternate end of the spectrum:
they are highly expressive recurrent neural networks representing the
best of breed predictor in multiple domains, with an enormous
parameter space, high training latency, low interpretability, and some
sensitivity to the specifics of the training regime.  We use a slight
variant of the `medium'-sized LSTM~\cite{zarembaSV14}.  We perform
comparisons for \lastfm, \brightkite, and \reuters; the \wiki dataset
is too small to showcase the strengths of LSTMs, so we omit it.
Table~\ref{tab:lstm} shows the results.
\begin{table} \label{lamp_lstm_same}
\centering
\small
\begin{tabular}{ l|ccc } 
\hline
Algorithm & \brightkite & \lastfm & \reuters \\
\hline
LAMP order 6, 1.5 iter & $38.4$ & $1054.6$ & $296.8$ \\
LSTM, short training time & $ 85.8$ & $1359.1$ & $105.4$ \\ 
LSTM, long training time & $ 51.0 $ & $525.7$ & $ 60.4 $ \\
\hline
\end{tabular}
\caption{Perplexity of LAMP and LSTM.  Short training time for LSTM is
  same as LAMP; long training time is  $20$x LAMP.}
\label{tab:lstm}
\end{table}

LAMP performs well on \brightkite\ relative to the LSTM.  Even after
20x training time the LSTM's perplexity is about 1/3 higher than LAMP,
at which point the LSTM begins to overfit.

On \lastfm, the LSTM with similar training time to LAMP performs about
30\% worse, but with additional training time, the LSTM performs
significantly better, attaining about 50\% of the test perplexity of
LAMP, reducing from about 10 bits to about 9 bits of uncertainty in
the prediction.  We hypothesize this is because of additional
structure within the music domain.

LSTMs are known to perform well on language data as in the
\reuters\ corpus, and pure sequential modeling is unlikely to capture
the nuance of language in the same way.  The results are consistent
with this expectation: LSTM with short training time already attains
around $1/3$ the test perplexity of LAMP, and with more training, the
LSTM improves to roughly $1/5$ of LAMP.
 
In summary, we expect that in general, LSTMs will outperform simpler
techniques for complex sequential tasks such as modeling language,
speech, etc.  We expect that LAMPs will be more appropriate in
settings in which Markov processes are typically used today: as
simple, interpretable, extensible sequence modeling techniques that
may easily be incorporated into more complex systems.  Nonetheless, it
is interesting that for datasets like \brightkite\ and \lastfm, LAMP
performs on par with LSTMs, indicating that LAMP models represent a
valuable new point on the complexity/accuracy tradeoff curve.



\newcommand{\CP}{{\cal P}}

\section{Discussion}
\label{sec:disc}

So far, we have formulated LAMP by the history distribution $w$ and a
transition matrix $P$, modeling temporal and contextual effects as
multiplicative. But while this model performs well, with more data
available, we might wish to relax this assumption.

The LAMP framework easily extends to a more general formulation in the
following manner.  Let $k$ be the support of $w$; we will assume $k$
is finite in this section.
\begin{defn}[Generalized LAMP]
Given a distribution $w$ on a finite support of size $k$ and
$\ell$ stochastic matrices
$\CP = \{ P^{(1)},\ldots,P^{(\ell)} \}$, and a function $f: [k] \rightarrow
[\ell]$, the \emph{Generalized Linear Additive Markov Process}
$\GLAMP(w, \CP, f)$ evolves according to the following 
transition rule:
\mymath{
\Pr[X_t = x_t \mid x_0, \ldots, x_{t-1}]
= \sum_{i=1}^{k} w_{i} P^{(f(i))}(x_{\max\{0,t-i\}}, x_t).
}
\end{defn}
This corresponds to a user transitioning from a previous state
dependent on the state's position in their history. In particular, the
temporal and contextual aspects are combined directly in the
stochastic matrices $P^{(i)}$, instead of only multiplicatively.  Note
that LAMP can be realized by making $\ell = 1$ and $f \equiv 1$, the
constant function.  Another interesting case is when we wish
to treat the state immediately prior to the current one with more
emphasis than those following.  In this case, $\ell = 2$, $f(1) = 1$,
and $f(i) = 2$, $i > 1$.

Despite this generalization, we can still extend some
of our results for LAMPs to GLAMPs. In particular, Theorem
\ref{equib-thm} holds.
\begin{thm} 
\label{thm:glamp}
$\GLAMP(w, \CP, f)$  has an equilibrium vector if and only if the 
matrix $P = \sum_{i = 1}^{k} w_i P^{(f(i))}$ is ergodic. Furthermore, 
this equilibrium vector is the same as equilibrium vector for 
the first-order Markov process induced by $P$. 
\end{thm}
Note that the matrices $P^{(i)}$ may not necessarily commute, so there is no simple characterization via the exponent process as
in the $\LAMP(w, P)$ case.  We prove from first
principles
(details omitted).

\section{Conclusions}
\label{sec:conc}

In this paper, we propose the linear additive Markov process, LAMP.  LAMP
incorporates history by utilizing a history distribution in
conjunction with a stochastic matrix.  While it is provably more
general than a first-order Markov process, it inherits many nice
properties of Markov processes, including succinct representation,
ergodicity, and quantifiable mixing time.  LAMPs can also be easily
learned.  Experiments validate that they go well beyond first-order
Markov processes in terms of likelihood.  Overall, LAMP is a powerful
alternative to a first-order Markov process with only negligible
additional cost in terms of space and parameter estimation.

There are several questions around LAMP that constitute interesting
future work.  They include obtaining more efficient and provably good
parameter estimation algorithms, carrying over standard Markov process
notions such as conductance, cover time, etc.\ to LAMP, and bringing
LAMP to other application domains such as NLP.  A particularly
intriguing question is if $\LAMP_2(w, P)$, which has just one more
parameter compared to a first-order Markov process, has a closed-form
parameter estimation.

\para{Acknowledgments}
The authors would like to thank Amr Ahmed and Jon Kleinberg for
fruitful discussions.

\balance
{\small
\bibliographystyle{abbrv}
\bibliography{lamp}
}

\end{document}